\documentclass{svproc}
\usepackage{graphicx} %use graph format
\usepackage{array}
\usepackage{threeparttable}
\usepackage{amsmath,amsfonts}
\usepackage{algorithmic}
\usepackage{algorithm}

\usepackage{multirow}
\usepackage{pgfplots}
\pgfplotsset{compat=1.18}
\usepackage{xcolor}
\usepackage{enumitem}

\DeclareMathOperator{\sgn}{sgn}

\definecolor{bblue}{HTML}{025497}
\definecolor{rred}{HTML}{C0504D}
\definecolor{ggreen}{HTML}{9BBB59}
\definecolor{ppurple}{HTML}{9F4C7C}
\definecolor{ggray}{HTML}{4D4D4D}

\usepackage{url}

\begin{document}
\mainmatter              % start of a contribution
\title{Lateral Velocity Model for Vehicle Parking Applications}
\titlerunning{Lateral Velocity Model}  % abbreviated title (for running head)
\author{Luis Diener\inst{1}\inst{2} \and Jens Kalkkuhl\inst{2} \and Markus Enzweiler \inst{1}}
\authorrunning{Luis Diener et al.} % abbreviated author list (for running head)
\tocauthor{Luis Diener, Jens Kalkkuhl and Markus Enzweiler}
\institute{Institute for Intelligent Systems, Esslingen University of Applied Sciences, Germany (ludiit01@hs-esslingen.de)\\
\and
Mercedes-Benz AG, Germany\\[0.5em]
\textit{This manuscript has been submitted to Vehicle System Dynamics for possible publication.}}

\maketitle              % typeset the title of the contribution

\begin{abstract}
Automated parking requires accurate localization for quick and precise maneuvering in tight spaces. While the longitudinal velocity can be measured using wheel encoders, the estimation of the lateral velocity remains a key challenge due to the absence of dedicated sensors in consumer-grade vehicles. Existing approaches often rely on simplified vehicle models, such as the zero-slip model, which assumes no lateral velocity at the rear axle. It is well established that this assumption does not hold during low-speed driving and researchers thus introduce additional heuristics to account for differences. In this work, we analyze real-world data from parking scenarios and identify a systematic deviation from the zero-slip assumption. We provide explanations for the observed effects and then propose a lateral velocity model that better captures the lateral dynamics of the vehicle during parking. The model improves estimation accuracy, while relying on only two parameters, making it well-suited for integration into consumer-grade applications.
\end{abstract}

\section{Introduction}
Automated parking has evolved from a basic driver assistance feature into a versatile system, able to perform parking maneuvers quickly and reliably even in complex and constrained environments. Unlike highway or urban driving, parking scenarios require centimeter-level precision due to the limited available space and the proximity of static and dynamic obstacles. The vehicle must follow the planned trajectory, making accurate egomotion and relative localization essential. Improved accuracy enables smaller safety margins and allows the system to park in tighter spaces.

\begin{figure}[ht]
\centering
\includegraphics[clip, trim=0.5cm 0.5cm 0.5cm 0.5cm,width=0.65\linewidth]{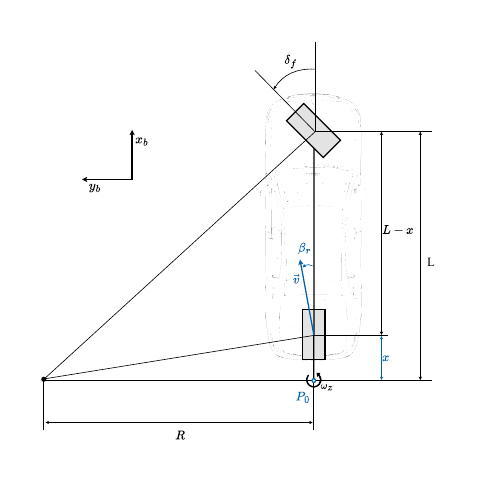}
\caption{Lateral velocity model resulting from a shift between the rear axis and the point $P_0$ where the zero-side-slip assumption holds.}
\label{fig:P0}
\end{figure}

Egomotion refers to the vehicle’s motion relative to an earth-fixed reference frame and is typically described by its 3D velocity, position, and attitude. In automotive systems, egomotion is estimated using proprioceptive sensors such as inertial measurement units (IMUs), wheel encoders, and steering angle sensors \cite{Marco.2020}. These sensors are reliable, provide high update rates, and meet automotive safety standards. However, accurate modeling of vehicle dynamics is additionally necessary to achieve the precision required for parking.

A key challenge in this context is the estimation of the lateral velocity $v_y$. While the longitudinal velocity $v_x$ can be measured using wheel encoders, there is no dedicated sensor for measuring $v_y$ in consumer-grade vehicles. As a result, lateral velocity is typically inferred using vehicle models, that serve as pseudo-measurements \cite{Marco.2020}. For consumer-grade vehicles, models with few parameters are preferred, as complex models increase uncertainty and require extensive vehicle-specific calibration.

In low-speed parking scenarios, where both velocity and lateral acceleration are small, the zero-slip model ($v_{y,r}=0$) is commonly applied. This model assumes zero side-slip at the rear axle. However, research shows that this assumption does not hold universally \cite{Brunker.2019}. The point of zero side-slip appears to shift depending on the vehicle type, configuration, and driving direction. Based on this observation, we propose a simple, yet effective, lateral velocity model for parking applications, that accounts for this variability while remaining suitable for consumer-grade applications.
{The main contributions are:}
\begin{itemize}
\item[$\bullet$] We introduce a simple lateral velocity model for vehicle parking applications, depicted in Fig.~\ref{fig:P0}.
\item[$\bullet$] We analyze and explain the lateral velocity deviations by demonstrating how known effects interact to produce the observed behavior.
\item[$\bullet$] We perform a parameter analysis of the developed model and evaluate how errors propagate under parameter uncertainty.
\end{itemize}

\section{Related Work}
We divide the current state of the art into three main categories, i.e. kinematic approaches, vehicle-dynamics models, and IMU dead-reckoning.

Kinematic models, either implemented as single-track model or two-track model, calculate motion from tire speeds. Using the tires' individual speeds and orientations, i.e. their steering angle, a dead-reckoning solution can be calculated \cite{Caltabiano.2004}, \cite{Kochem.2002}. This base assumption ignores wheel-slip in both longitudinal and lateral direction, causing trajectory deviations over time. Furthermore, this approach requires exact knowledge about and calibration of the individual wheel steering angles.

Vehicle-dynamics approaches are also based on the single or two-track model, but estimate motion based on tire forces, rather than kinematics. Tire forces can be modeled using a range of approaches, from linear slip-force gradients \cite{Bai.2012}, over empirical tire models \cite{Pacejka.2012}, and up to full physical simulations \cite{Gallrein.2014}, \cite{Oertel.2012}. While linear tire models provide good results for normal driving, they seem to fall short during parking maneuvers. On the other hand, complex models always require additional parameters, that can be hard to obtain and are often uncertain themselves.

IMU dead reckoning relies on double-integration of angular rates and accelerations to estimate the position. Considering the accuracy of automotive-grade IMUs this method yields poor results without stabilization. However, velocity measurements from wheel encoders can be used to stabilize the solution \cite{Klier.2008}, providing a better estimate of the position. Since the lateral velocity cannot be measured in consumer-grade vehicles, additional model assumptions are required \cite{Marco.2020}.

Solutions in practice usually rely on a combination of the aforementioned approaches to increase accuracy and robustness \cite{Han.2020}. Brunker et al. \cite{Brunker.2019} fuse the base variants within an information filter and switch between models to achieve better accuracy. Moreover, they use a corrected side-slip angle from ground-truth-generated look-up tables. With their approach, they reliably achieve localization errors below 20 cm for parking maneuvers. These typically include parallel and perpendicular maneuvers with an average length of 10 m to 20 m.
Han et al. \cite{Han.2025} developed a kinematic model and perform offline optimization using ground-truth information and simulations. They optimize the contribution of individual wheels depending on the driving situation.
Other approaches try to extend the zero-side-slip assumption by assigning it a driving-situation-dependent uncertainty. This is either done using vehicle modeling \cite{Mori.2025} or data-driven optimization \cite{Brossard.2019}.
Further solutions employ constraints on the vehicle model \cite{Du.2023} and fuse additional sensors such as GNSS \cite{Fazekas.2020}, camera \cite{Liang.2022}, \cite{Bev.2016} and radar \cite{Diener.2024}, \cite{Diener.2025}.

In summary, there exist three main variants to perform wheel-encoder-based localization for automated parking. Using a combination of these models; data-driven optimization; or by applying additional heuristics, researchers developed localization algorithms with increased robustness and accuracy.
Two gaps can be identified in the aforementioned literature. First, there appears to be a lack of focus on solutions that are applicable to consumer-grade vehicles. In such settings, individual vehicles cannot be calibrated as extensively as is often required. Moreover, the calibration of vehicle lines and variants quickly becomes time and cost intensive if special maneuvers and ground-truth measurement systems are required.
Second, the observed deviations between models and real-world data is rarely discussed and analyzed. While minor differences should be expected, the observed deviations hint towards a gap in the current understanding and application of low-speed vehicle dynamics.

In this paper, we first explain the underlying effects and their interaction that cause the observed behavior (Sec. 3). We then provide a simple model to consider these deviations, that is suited for consumer-grade passenger vehicles (Sec. 4). In addition, we perform a parameter perturbation analysis, allowing for the derivation of performance requirements (Sec. 5). Finally, we show experimental results in Sec. 6. Tab.~\ref{tab:glossary} lists values and parameters used within the scope of the paper.

\renewcommand{\arraystretch}{1.2}
\begin{table}[ht]
\centering
\begin{threeparttable}
\caption{Glossary.}
\label{tab:glossary}
\begin{tabular}{wc{0.1\linewidth}|wl{0.75\linewidth}}
\hline
\textbf{Symbol} & \textbf{Description} \\
\hline
$v_x$ & Longitudinal velocity [m/s] \\
$v_y$ & Lateral velocity [m/s] \\
$v_{y,r}$ & Lateral velocity at the rear axle \\
$\omega_z$ & Yaw rate [rad/s] \\
$f_y$ & Lateral specific force [m/s$^2$] \\
%$\beta$ & Side-slip angle [rad] \\
$\beta_r$ & Side-slip angle at the rear axle [rad] \\
$\delta_f$ & Front steering angle [rad] \\
$F_i$ & Force acting at position $i$ [N] \\
$M_z$ & Torque acting around the yaw axis of the vehicle [Nm] \\
$\Delta\delta_A$ & Ackermann deviation [rad]\\
\hline
$m$ & Vehicle mass [kg]\\
$c_{f/r}$ & Front/rear lateral tire stiffness [N/rad]\\
%$c_r$ & Combined rear lateral tire stiffness [N/rad]\\
$\rho_{sg}$ & Side-slip gradient [rad s$^2$/m]\\
$J_zz$ & Yaw inertia [kgm$^2$]\\
$S_{Hy\varphi}$ & Horizontal shift in the MF tire model [rad]\\
$K_{yR\varphi 0}$ & camber stiffness [N/rad]\\
$K_{y\alpha 0}$ & lateral tire stiffness [N/rad] \\
$R_0$ & unloaded tire radius [m]
\end{tabular}
\end{threeparttable}
\end{table}

\section{Low-Speed Driving Behavior}
\subsection{Observed Behavior}
 When evaluating ego-motion data from parking maneuvers it can be observed that the zero-slip assumption at the rear axis does not hold. Brunker et al. \cite{Brunker.2019} also mentioned this discrepancy and modeled it using a look-up table linking the front steering angle with additive side slip. They used separate models for forward and reverse driving.

\begin{figure}[ht]
\centering
\includegraphics[width=0.8\linewidth]{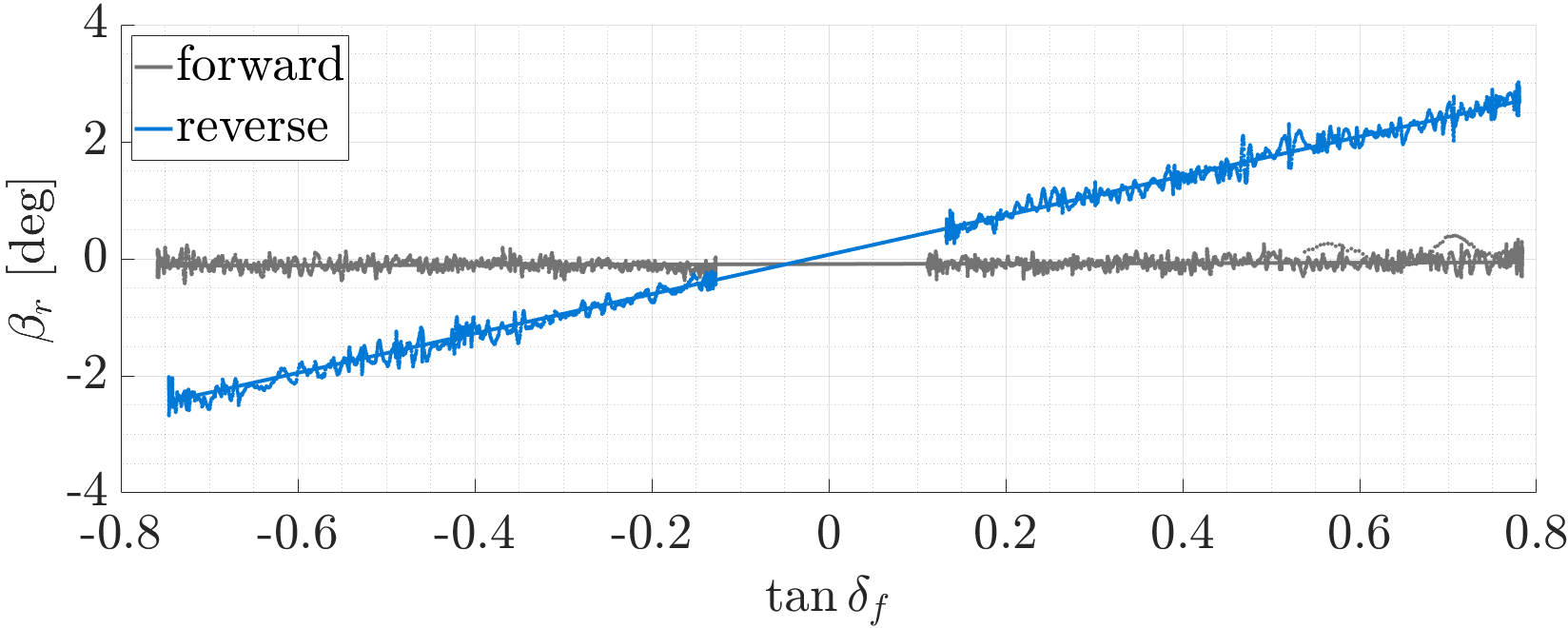}
\caption{Evaluated side-slip behavior during low-speed maneuvering with fitted lines for forward (blue) and reverse driving (gray). The vehicle is driving a dedicated calibration maneuver to obtain these clean results.}
\label{fig:behavior}
\end{figure}

Fig. \ref{fig:behavior} shows the side-slip angle at the rear axis $\beta_r$ of a vehicle performing low-speed calibration maneuvers. The ground-truth system to obtain these measurements is an inertial navigation system with real-time-kinematics information allowing for centimeter-level accuracy. We plot the side-slip angle $\beta_r$ over the tangent of the front-wheel steering angle $\tan\delta_f$. The behavior seems to be direction dependent and the relationship appears to be linear. While in this particular example there is no effect for forward driving, a significant side-slip angle appears during reverse driving. 

The lateral acceleration during parking maneuvers is relatively small ($<$1 m/s$^2$) and thus the forces required to keep the vehicle within its turn are also small, which, according to the single-track model, should lead to small slip angles. Several researchers investigated dependencies between tire behavior and speed. Guo et al. \cite{Guo.2005} analyze the effect of tire rolling speed on various tire parameters, but mainly found effects on steering torque. Similarly, Garcia et al. \cite{Garcia.2014} developed a model for lower speeds but found no noticeable speed-dependent effects. Wei et al. \cite{Wei.2016} and Cao et al. \cite{Cao.2019} also found and modeled speed-dependent torque effects with further findings concerning rolling resistance. Lugaro et al.\cite{Lugaro.2016} investigated the turn-slip effect to model tire forces for parking maneuvers. Complemented by later findings from Besselink et al. \cite{Besselink.2022}, turn slip appears to have a significant effect at low speeds, altering the tire forces. Brunker et al. \cite{Brunker.2019} investigate Ackermann deviations, that cause additional forces impacting the behavior at large steering angles.

\subsection{Turn-Slip Effect at Low Speeds}
Literature hints towards turn slip $\varphi_{ij}$ having a significant impact on low-speed vehicle dynamics. Turn slip defines the ratio between yaw rate $\omega_z$ and the velocity $v_{ij}$ in the tire contact patch:
\begin{align}
\varphi_{ij} = -\frac{\omega_z}{v_{ij}}
\end{align}
Considering the speeds during normal driving ($v>10$ m/s), this ratio is usually so small that the effect becomes negligible. However, parking maneuvers are characterized by low speeds and high yaw rates increasing the effect of turn slip on the tire behavior \cite{Besselink.2022}. 

Turn slip occurs due to varying slip gradients across the tire contact patch during low-speed driving. Normally, the slip velocity remains uniform across the contact area. However, in low-speed turns a lateral velocity gradient along the tire patch is introduced, causing the contact patch to experience varying local slip conditions. This variation then leads to an altered force transmission of the tire.
Turn slip is described in the well-known Pacejka Magic Formula (MF) tire model \cite{Pacejka.2012}. There, it causes a horizontal shift $S_{Hy\varphi}$ of the slip-force function. For small turn-slip ratios (which is the case for regular parking maneuvers) the relationship remains linear:
\begin{align}
S_{Hy\varphi}=K_{yR\varphi 0}/K_{y\alpha 0}R_0\varphi\sgn(v_{ij}),
\end{align}
with the camber stiffness $K_{yR\varphi 0}$, the lateral tire stiffness $K_{y\alpha 0}$ and the unloaded tire radius $R_0$. Moreover, the effect is dependent on the driving direction, indicated by the sign function, that is applied to the tire velocity $\sgn(v_{ij})$.
Given the single-track model relationship
\begin{align}
\varphi = \dfrac{\omega_z}{v_x} = \dfrac{\tan\delta_f}{L},
\end{align}
we receive a linear relationship between the horizontal shift $S_{Hy\varphi}$ and the tangent of the front steering angle $\tan\delta_f$. This essentially acts as an additive component to the rear side-slip angle $\beta_r$ that is roughly characterized by:
\begin{align}
\Delta\beta_r \approx \sgn(v_x)\underbrace{\frac{K_{yR\varphi 0}}{K_{y\alpha 0}}\frac{R_0}{L}}_{\text{const.}}\tan\delta_f
\end{align}
Thus, we already have a model that describes a driving-direction-dependent and linear relationship between the rear side-slip angle $\beta_r$ and the tangent of the steering angle $\tan\delta_f$.

\subsection{Ackermann Deviation}
The Ackermann steering geometry is usually desired for consumer-grade vehicles. With this geometry, each tire follows the path around the same center of rotation, minimizing tire forces. To achieve this, the inside wheel requires a greater steering angle than the outer wheel. In a left turn, the relationship becomes
\begin{align}
\tan(\delta_{fl})=\dfrac{L}{R-b_f/2},\quad\tan(\delta_{fr})=\dfrac{L}{R+b_f/2}
\end{align}
While this geometry is aimed for, consumer-grade vehicles are usually designed with less difference in steering angle than required. This is done both due to space constraints and to enforce under-steering behavior \cite{Simione.2002}. Thus, an Ackermann deviation $\Delta\delta_A<0$ is induced.

\begin{figure}[ht]
\centering
\includegraphics[clip, trim=0.7cm 0.5cm 1cm 0.5cm,width=0.65\linewidth]{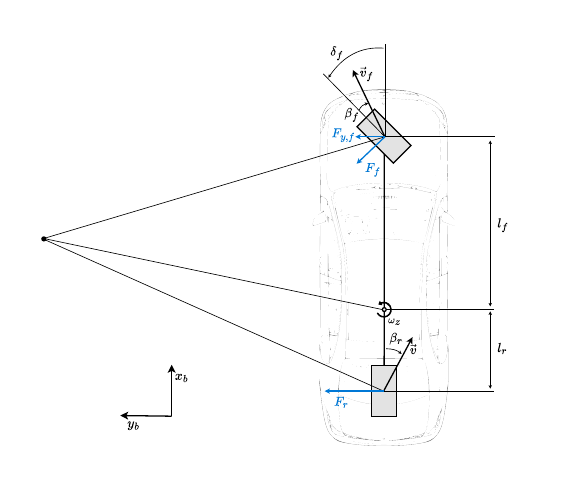}
\caption{Single-track model with side-slip angle $\beta$, velocity vector $\vec{v}$, yaw rate $\omega_z$, lever arms to the front $l_f$ and rear $l_r$, slip-angles $\beta_i$, steering angle $\delta_f$ and tire forces $F_{y,i}$}
\label{fig:stm}
\end{figure}

To analyze the effect of Ackermann deviations on the side-slip behavior, we employ the single-track model. We introduce both a disturbance force $\Delta F_y$ and disturbance torque $\Delta M_z$ into the single-track model. Applying the force and moment equilibrium to the single-track model in Fig. \ref{fig:stm} then leads to 
\begin{align}
\label{eq:stm}
m a_y &= F_{y,f}+F_{y,r}+\Delta F_y\\
J_{zz} \dot{\omega}_{z} &= F_{y,f}l_f-F_{y,r}l_r + \Delta M_z
\label{eq:stm2}
\end{align}
with the yaw inertia $J_{zz}$, the vehicle mass $m$, the lever arms $l_f$ and $l_r$ as well as the disturbance force $\Delta F_y$ and torque $\Delta M_z$.
We eliminate $F_{y,f}$ by inserting (\ref{eq:stm2}) into (\ref{eq:stm}). This leads to
\begin{align}
\label{eq:stm_big}
ma_y = \dfrac{J_{zz}\dot{\omega}_z}{l_f}+\dfrac{F_{y,r}l_r}{l_f}+F_{y,r} + \Delta F_y - \dfrac{\Delta M_z}{l_f}
\end{align}
For normal driving $(|a_y|<4m/s^2)$ the tire behavior can be modeled using the linear relationship
\begin{align}
\label{eq:fyr}
F_{y,r} = c_r\beta_r.
\end{align}
and the side-slip gradient
\begin{align}
\label{eq:sg_def}
\rho_{sg} = \dfrac{m l_f}{c_r (l_f+l_r)}
\end{align}
Inserting Eq. \eqref{eq:fyr} into Eq. (\ref{eq:stm_big}) allows us to derive
\begin{align}
\label{eq:stm_ultra}
ma_y =& \dfrac{J_{zz}\dot{\omega}_z}{l_f}+c_r\beta_r\left(\dfrac{l_r}{l_f}+1\right)+ \Delta F_y-\dfrac{\Delta M_z}{l_f}
\end{align}
Since we assume steady-state behavior $(\dot{\omega}_z=0)$, we discard the corresponding term in Eq. (\ref{eq:stm_ultra}). We rearrange Eq. (\ref{eq:sg_def}) and insert it into Eq. (\ref{eq:stm_ultra}):
\begin{align}
\label{eq:vy_1}
\beta_r = - \rho_{sg}  a_y + \dfrac{\rho_{sg}}{m} \left(\Delta F_y - \dfrac{\Delta M_z}{l_f}\right)
\end{align}

\begin{figure}[ht]
\centering
\includegraphics[width=0.6\linewidth]{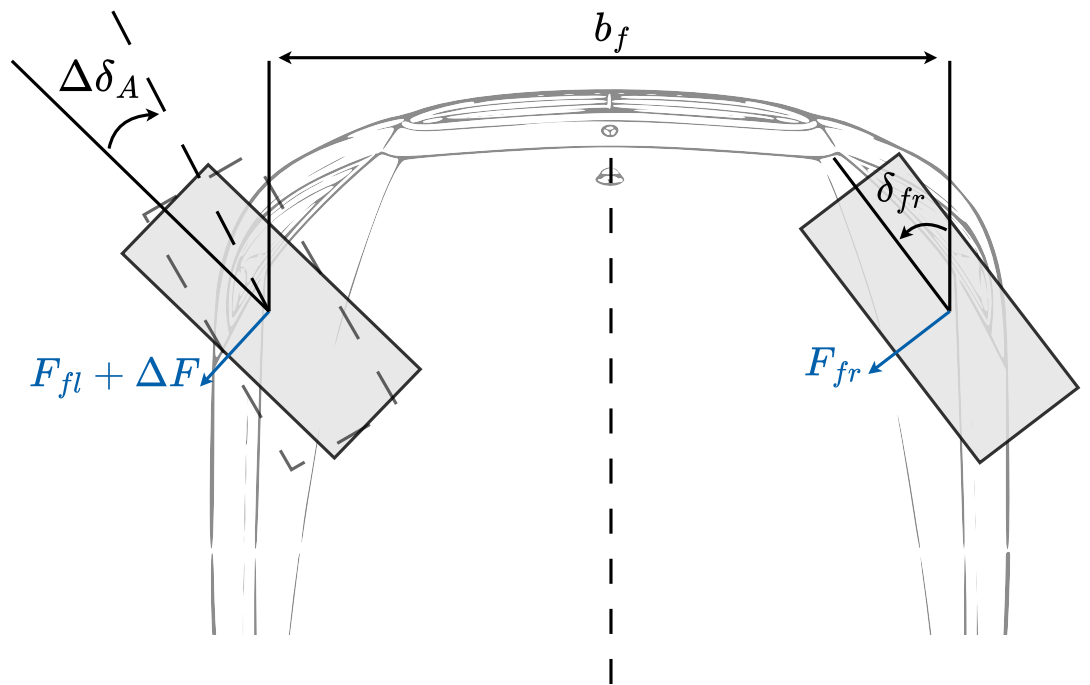}
\caption{Two-track model at the front axis, where additional side-slip and thus tire force $\Delta F$ is being generated. This is caused by the deviation from Ackermann steering, that would minimize the slip angle at each tire.}
\label{fig:stm_ackermann}
\end{figure}
We approximate the disturbance force and torque using the two-track model at the front axis, as depicted in Fig. \ref{fig:stm_ackermann}. Given Ackermann steering, the two-track model behaves similarly to the single-track model. Thus, only small slip angles occur at each tire, keeping the vehicle within the turn. If Ackermann deviations occur, the slip angles at the front axis consequently become larger, leading to increased tire forces. From Fig. \ref{fig:stm_ackermann} we derive for the disturbance force $\Delta F_y$ and torque $\Delta M_z$:
\begin{align}
\Delta F_y &= \cos\delta_{f}\Delta F\\
\Delta M_z &= l_f\cos\delta_f\Delta F + \frac{b_f}{2}\sin\delta_f\Delta F,
\end{align}
where we approximate 
\begin{align}
\Delta F \approx \frac{c_f}{2}\Delta\delta_A
\end{align}
Inserting into Eq. (\ref{eq:vy_1}) leads to 
\begin{align}
\label{eq:model}
\beta_r &=- \rho_{sg}  a_y + \dfrac{\rho_{sg}b_f}{4\,l_f m}\sin\delta_fc_f\Delta\delta_A
\end{align}
For our particular vehicle the relationship between the inside-wheel steering angle and the Ackermann deviation is mostly linear. In that case, the relationship in Eq. (\ref{eq:model}) behaves quadratic for small steering angles.

\begin{figure}[ht]
\centering
\includegraphics[width=0.8\linewidth]{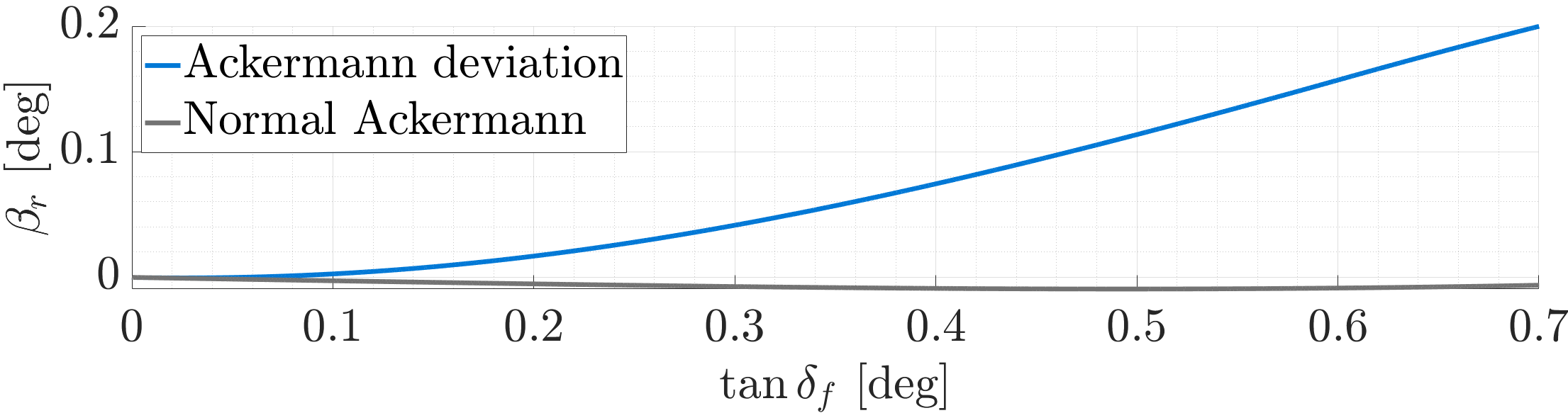}
\caption{Effect of Ackermann deviation on side-slip angle (red). The normal Ackermann behavior is depicted as well (blue).}
\label{fig:a_plot}
\end{figure}

We simulate the Ackermann effect using a non-linear two-track model \cite{Vietinghoff.2007} with a Pacejka MF tire model excluding the tire-spin effect \cite{Pacejka.2012}. Fig. \ref{fig:a_plot} depicts the results for both normal Ackermann steering and the Ackermann deviation variant. Importantly, this effect is independent of the driving direction. Thus, the side-slip at the rear axis changes depending on the steering angle and the Ackermann deviation only.

For consumer-grade vehicles, where $\Delta\delta_A<0$, the effect of Ackermann deviation generates a negative disturbance torque, that causes a positive shift in side-slip angle, regardless of the driving direction.

\begin{figure}[ht]
\centering
\includegraphics[width=0.8\linewidth]{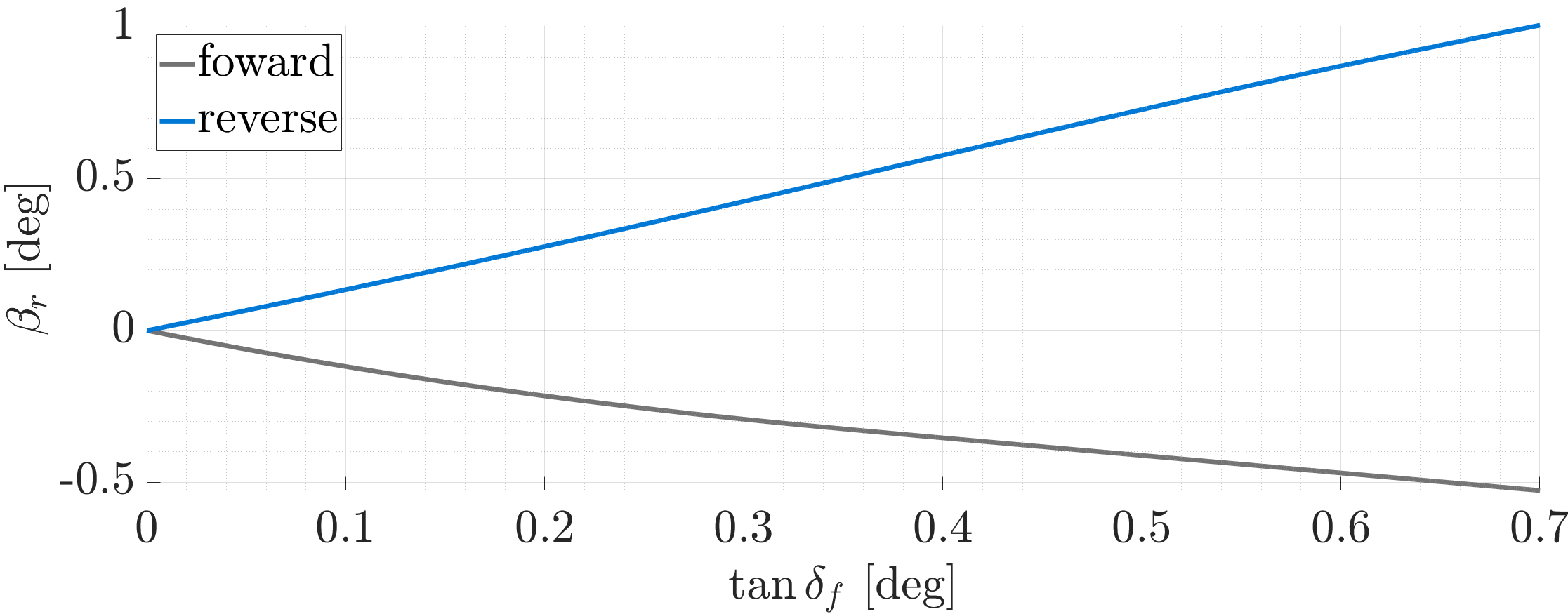}
\caption{Simulated effect of combined Ackermann deviation and turn slip for forward and reverse driving.}
\label{fig:full_effect}
\end{figure}

Combining the effect of Ackermann deviation with the tire-spin effect leads to the observed behavior. Simulating both effects results in the behavior shown in Fig. \ref{fig:full_effect}, which coincides with the observations made on real data. The turn-slip effect causes additional side slip, with its sign depending on the driving direction. The asymmetry between forward and reverse driving is caused by deviations from Ackermann steering. There, an additive component acts on the side slip, independent of the driving direction. Thus, the turn-slip effect is amplified during reverse driving and decreased during forward driving maneuvers.

\subsection{Other Effects}
In analyzing the observed behavior, we investigated several other potential effects to explain the deviations.

Tires are not the only source of elasticity when forces act on the vehicle as additional compliance arises from the wheel suspension, steering linkage, and chassis. We analyzed the elastokinematic behavior of both the front and rear axles and found no significant influence during parking maneuvers. The forces involved are too small to produce notable compliance, with maximum toe deviations reaching only $0.1^\circ$. Moreover, we excluded any meaningful kinematic contribution from the suspension geometry, as the variation in wheel toe across the deflection range is minimal. We also evaluated potential disturbance torques from the drivetrain and frictional forces but found them insufficient to generate the side-slip angles in question. 

Additional effects include wheel-load transfers in combination with local friction coefficients that can also influence the lateral vehicle dynamics. During low-speed maneuvering the tire exhibits large camber angles, that also contribute to a changed behavior. We conclude that, while turn slip appears to be the main driver of the observed low-speed behavior, there are many additional effects contributing to the final model. This is also why modeling this behavior is questionable, especially in a consumer-grade environment. Each modeled effect introduces additional vehicle and tire-specific parameters, increasing the calibration effort and the overall uncertainty.

\section{Lateral Velocity Model}
\subsection{Model Description}
Our results demonstrate that passenger vehicles can exhibit considerable side-slip angles at low speeds. The magnitude of this low-speed effect is influenced by multiple factors, most notably vehicle geometry and tire properties. However, in a consumer-grade context each additional parameter contributes to the overall calibration effort. As a result, entire vehicle lines, including their sub-variants, would require dedicated parameter sets to account for these influences. Brunker et al. \cite{Brunker.2019} used a single-track model with fifth-order polynomials to capture the side-slip angle deviations. To mitigate this complexity, we desire a simplified model that captures the observed behavior while relying on only a minimal number of parameters.

Using data from several different vehicles, we observe that the relationship between the tangent of the front steering angle $\delta_{f}$ and the rear side-slip angle $\beta_r$ appears to behave linearly, as we already depicted in Fig. \ref{fig:behavior}.  Thus, we model the relationship:
\begin{align}
\label{eq:old_model}
\beta_r = -k_{\rho}\tan\delta_f,
\end{align}
where $k_{\rho}$ is a newly defined parameter. This $\delta$-$\beta$ model proves to be impracticable for two main reasons. First, it depends on the steering angle signal that is usually uncalibrated. Second, it requires the measurement of the side-slip angle, that becomes ill-defined at low speeds and is hard to measure even for the ground-truth system. Thus, estimating the parameter $k_{\rho}$, even during vehicle testing, becomes a challenge.

We therefore propose a different model, the $\omega_z$-$v_y$ model, that covers that same behavior but is more robust during parameter estimation and independent of the steering angle:
\begin{align}
\label{eq:new_model}
v_{y,r} = -x_{\rho}\, \omega_z,
\end{align}
where $x_\rho$ corresponds to the lever arm between the rear axis and a point $P_0$ on the longitudinal axis of the vehicle, at which the zero side-slip assumption is valid. For a vehicle in a steady-state turn, the kinematic bicycle model offers the turn radius $R$ as a function:
\begin{align}
R=\dfrac{v_x}{\omega_z}
\end{align}
Moreover, it can also be seen for the front-wheel angle:
\begin{align}
\tan\delta_f=\dfrac{L}{R}
\end{align}
Since we assume that the point $P_0$, where the lateral velocity is zero, does not coincide with the center of the rear axis anymore, we now obtain:
\begin{align}
\tan\delta_f=\dfrac{L-x_{\rho}}{v_x}\omega_z
\end{align}
and
\begin{align}
v_{y,r} = -x_{\rho}\, \omega_z,
\end{align}
Combining those two equations will result in:
\begin{align}
\dfrac{v_{y,r}}{v_x}=\beta_r=-\dfrac{x}{L-x_{\rho}}\tan\delta_f,
\end{align}
where 
\begin{align}
\dfrac{x}{L-x}=k_{\rho}
\end{align}
is our previously defined parameter. For the vehicle depicted in Fig. \ref{fig:behavior}, we estimated the parameter $k_{\rho}=0.054$, which corresponds to $x_\rho =-21.0$ cm. Thus, during a reverse parking maneuver, the point $P_0$, where the zero side-slip condition holds, is actually 21.0 cm behind the rear axis.

Using this new model now only requires a lateral velocity as well as a calibrated yaw rate, which is easier to obtain from the ground-truth system. Moreover, we translated the previously abstract parameter into a physically meaningful distance.

\subsection{Parameter Estimation}
To motivate the usage of the new $\omega_z$-$v_y$ model (Eq.~(\ref{eq:new_model})) instead of relying on the $\delta$-$\beta$ model (Eq.~\eqref{eq:old_model}), we compare the estimation of the parameter using data from regular parking maneuvers. While the initial data from Fig. \ref{fig:behavior} depicts dedicated calibration maneuvers, we now estimate the parameter using data captured in a parking lot, while also performing parking maneuvers. This introduces disturbances and noise, making parameter estimation more challenging.

\begin{figure}[ht]
\centering
\includegraphics[width=0.7\linewidth]{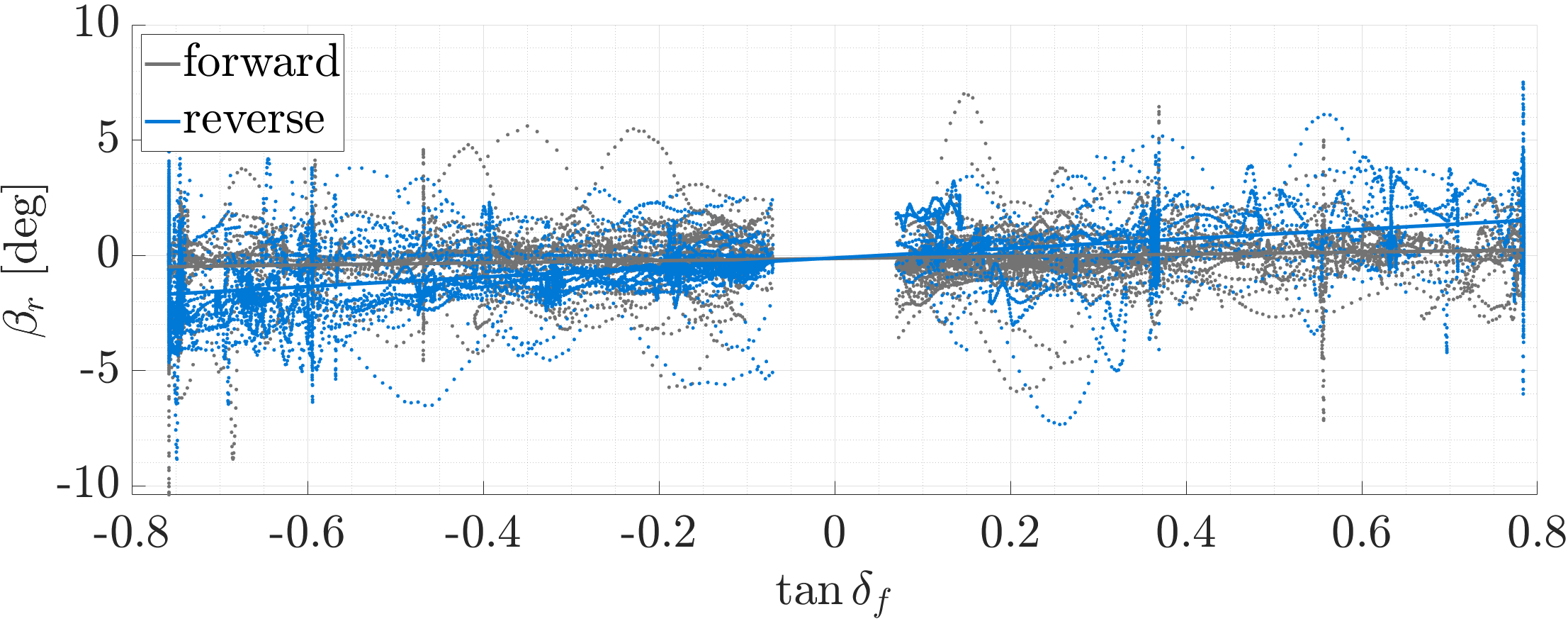}
\includegraphics[width=0.7\linewidth]{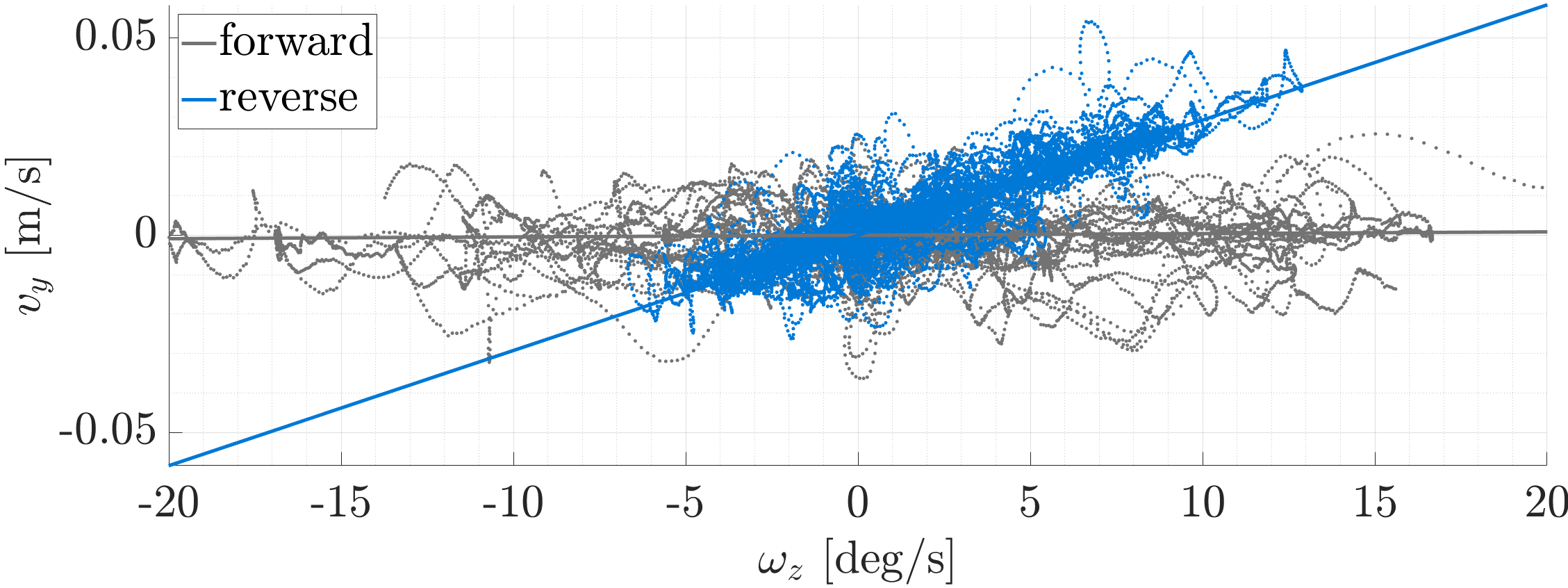}
\caption{Comparison between the parameter estimation using the model based on side-slip angle and the new model based on lateral velocity.}
\label{fig:comparison}
\end{figure}

Fig. \ref{fig:comparison} shows the regression of both the $\delta$-$\beta$ model and the $\omega_z$-$v_y$ model for the same parking maneuvers. The data is noisy in both cases, but the regression correlates better in the $\omega_z$-$v_y$ model variant. Moreover, Tab.~\ref{tab:comparison} lists the estimated parameters $k_{\rho}$ and $x_{\rho}$ in comparison with the ground-truth values. There, we see that the $\omega_z$-$v_y$ model performs better, leading to an estimation error of less than 1 cm. On the other hand, the $\delta$-$\beta$ model misses the forward and reverse parameters by 2.3 cm and 7.3 cm respectively. The ground truth was obtained using calibration maneuvers, where both models lead to the same parameters.

\renewcommand{\arraystretch}{1.3}
\begin{table}[ht]
\centering
\begin{threeparttable}
\caption{Comparison of lateral velocity models for parameter estimation. Depicted are the deviations from ground truth.}
\label{tab:comparison}
\begin{tabular}{wl{2cm}|wc{2cm}wc{2cm}|wc{2cm}wc{2cm}}
%\hline
 & \multicolumn{2}{c|}{\textbf{$\delta$-$\beta$ model}} & \multicolumn{2}{c}{\textbf{$\omega_z$-$v_y$ model}}\\
\hline
&$k_{\rho}$ & $x_{\rho}$&$k_{\rho}$ & $x_{\rho}$\\
\hline
forward & 0.8\% & 2.3 cm & $\mathbf{0.1\%}$ & $\mathbf{0.1}$ cm\\
reverse & $-2.4$\% & $-7.3$ cm & $\mathbf{-0.4\%}$ & $\mathbf{-0.9}$ cm
\end{tabular}
\begin{tablenotes}
\item Note: Depicted are the deviations of both parameters from their ground-truth values. The $\omega_z$-$v_y$ model achieves the best performance.
\end{tablenotes}
\end{threeparttable}
\end{table}

\section{Parameter Disturbance Analysis}
\begin{figure}[ht]
    \centering
	    \includegraphics[width=0.6\linewidth]{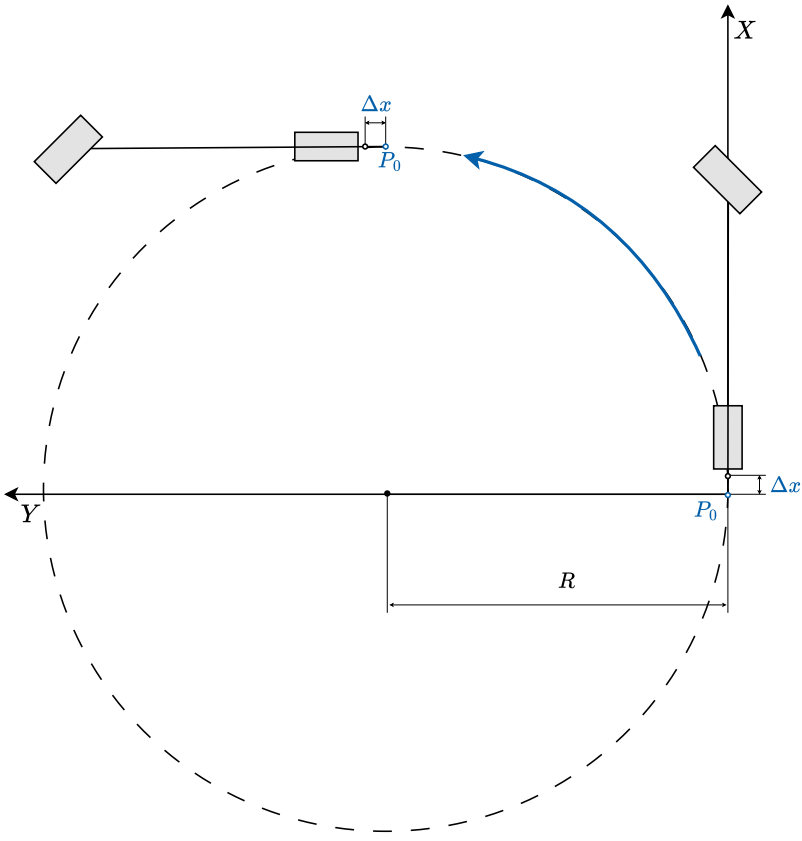}
    \caption{Driving a steady state circle.}
    \label{fig:circle}
\end{figure}

We showed that the combined turn slip and Ackermann deviation can be interpreted as a physical distance $x_{\rho}$ between the rear axis and a point $P_0$, where the zero side-slip assumption is valid. We now investigate the influence of a parameter error $\Delta x$ on the estimation of the relative vehicle position. This way we can quantify the impact of the parameter and its uncertainties on localization performance. The following assumptions are made:
\begin{itemize}
    \item[$\bullet$] A full steady-state circle is driven on a horizontal plane, as shown in Fig.~\ref{fig:circle}.
    \item[$\bullet$] The point $P_0$, where the lateral velocity is zero in reality, is marked in blue.
    \item[$\bullet$] The point where the lateral velocity is nominally assumed to be zero has a distance of $\Delta x$ from $P_0$.
\end{itemize}

The differential equations describing the planar motion are given by
\begin{align}
    \dot{\Psi} &= \omega_z \label{eq:psi_dot} \\
    \dot{X} &= v_x \cos\Psi - v_y \sin\Psi \label{eq:x_dot} \\
    \dot{Y} &= v_x \sin\Psi + v_y \cos\Psi \label{eq:y_dot}
\end{align}
Under the condition
\begin{equation}
    \omega_z = \text{const.}, \quad \Psi(0) = 0 \label{eq:init_psi}
\end{equation}
the position can be described as
\begin{align}
    \Psi(t) &= \omega_z t \label{eq:psi_t} \\
    \dot{X} &= v_x \cos(\omega_z t) - v_y \sin(\omega_z t) \label{eq:x_dot_param} \\
    \dot{Y} &= v_x \sin(\omega_z t) + v_y \cos(\omega_z t) \label{eq:y_dot_param}
\end{align}
The initial conditions for the undisturbed solution are as shown in Fig.~\ref{fig:P0}:
\begin{align}
    X(0) &= 0, \quad Y(0) = 0 \label{eq:init_pos} \\
    v_x &= \text{const.}, \quad v_y \equiv 0 \label{eq:vx_vy_const}
\end{align}
This results in the simplified equations:
\begin{align}
    \dot{X} &= v_x \cos(\omega_z t) \label{eq:x_dot_simple} \\
    \dot{Y} &= v_x \sin(\omega_z t) \label{eq:y_dot_simple}
\end{align}
which have the general solution:
\begin{align}
    X(t) &= \frac{v_x}{\omega_z} \sin(\omega_z t) + C_1 \label{eq:x_gen} \\
    Y(t) &= -\frac{v_x}{\omega_z} \cos(\omega_z t) + C_2 \label{eq:y_gen}
\end{align}
Taking into account the initial condition~\eqref{eq:init_pos}, we arrive at the undisturbed solution:
\begin{align}
    X(t) &= \frac{v_x}{\omega_z} \sin(\omega_z t) \label{eq:x_sol} \\
    Y(t) &= \frac{v_x}{\omega_z} (1 - \cos(\omega_z t)) \label{eq:y_sol}
\end{align}
As shown in Fig.~\ref{fig:circle}, the point $P_0$ moves on a circle with a radius
\begin{equation}
    R = \frac{v_x}{\omega_z} \label{eq:radius}
\end{equation}
Next, we compute the solution under the assumption that the zero-slip assumption $v_y \equiv 0$ does not hold. Instead, the lateral velocity is perturbed by
\begin{equation}
    v_y = \Delta x \, \omega_z \label{eq:vy_perturb}
\end{equation}
The perturbed differential equations are
\begin{align}
    \dot{X}' &= v_x \cos(\omega_z t) - \Delta x \, \omega_z \sin(\omega_z t) \label{eq:x_dot_pert} \\
    \dot{Y}' &= v_x \sin(\omega_z t) + \Delta x \, \omega_z \cos(\omega_z t) \label{eq:y_dot_pert}
\end{align}
and have the general solution:
\begin{align}
    X'(t) &= \frac{v_x}{\omega_z} \sin(\omega_z t) + \Delta x \cos(\omega_z t) + C_1 \label{eq:x_pert_gen} \\
    Y'(t) &= -\frac{v_x}{\omega_z} \cos(\omega_z t) + \Delta x \sin(\omega_z t) + C_2 \label{eq:y_pert_gen}
\end{align}
With the initial conditions
\begin{equation}
    X'(0) = 0, \quad Y'(0) = 0 \label{eq:init_pert}
\end{equation}
we obtain the perturbed solution:
\begin{align}
    X'(t) &= \frac{v_x}{\omega_z} \sin(\omega_z t) + \Delta x (\cos(\omega_z t) - 1) \label{eq:x_pert_sol} \\
    Y'(t) &= \frac{v_x}{\omega_z} (1 - \cos(\omega_z t)) + \Delta x \sin(\omega_z t) \label{eq:y_pert_sol}
\end{align}
Taking into account that Eq.~\eqref{eq:x_pert_sol} contains the undisturbed solution~\eqref{eq:x_sol}, the perturbation of the position caused by the parameter error $\Delta x$ can be derived as:
\begin{align}
    \Delta X(t) &= X'(t) - X(t) = \Delta x (\cos(\omega_z t) - 1) \label{eq:delta_x} \\
    \Delta Y(t) &= Y'(t) - Y(t) = \Delta x \sin(\omega_z t) \label{eq:delta_y}
\end{align}
From the above equations, the following consequences arise:
\begin{itemize}
    \item[$\bullet$] The overall positioning error on a steady-state circle is given by
    \begin{equation}
        e = \sqrt{\Delta X^2 + \Delta Y^2} = \Delta x \sqrt{2 (1 - \cos(\omega_z t))} \label{eq:error_general}
    \end{equation}
    \item[$\bullet$] The maximum position error is obtained after a $180^\circ$ turn:
    \begin{equation}
        e_{\text{max}} = 2 \Delta x \label{eq:error_max}
    \end{equation}
    \item[$\bullet$] The error after a $90^\circ$ turn is
    \begin{align}
        \Delta X &= -\Delta x \label{eq:delta_x_90} \\
        \Delta Y &= \Delta x \label{eq:delta_y_90} \\
        e_{90} &= \sqrt{2} \, \Delta x \label{eq:error_90}
    \end{align}
\end{itemize}
The latter will be the approximate positioning error in a $90^\circ$ parking maneuver. For our particular example with $x_{\rho,r}=-21$ cm, this would result in a positioning error of $e_{90}=29.7$ cm for a reverse perpendicular parking maneuver. For a forward parking maneuver the error would be close to zero. 

In summary, the effect we described has significant impact on the localization accuracy during parking, causing localization errors outside the required accuracy. Furthermore, this analysis allows us to derive requirements for the accuracy of the parameter itself.

\section{Experimental Results}
\subsection{Localization Filter}
We apply our lateral velocity model within a localization filter to validate improved localization performance.
The motion model describes the vehicle's body-fixed velocity $\mathbf{v}$, its position $\mathbf{p}$, and its attitude $\mathbf{q}$, represented as a Hamilton quaternion. The derivatives of these states are expressed as \cite{Sola.2017}
\begin{align}
\label{eq:vmo_1}
\dot{\mathbf{v}} &= \mathbf{a}+\mathbf{R}({\mathbf{q}})^\top\mathbf{g}-[\boldsymbol{\omega}\times] \boldsymbol{v}\\
\dot{\mathbf{q}} &= \boldsymbol{\omega}\label{eq:vmo_2} \\
\dot{\boldsymbol{p}} &= \mathbf{R}({\mathbf{q}})\boldsymbol{v} \label{eq:vmo_3}
\end{align}
with $\mathbf{g}=\begin{bmatrix} 0&0&9.81\text{m}/\text{s}^2 \end{bmatrix}^\top$, the direction cosine matrix $\mathbf{R}({\mathbf{q}})$, the measured acceleration vector $\mathbf{a}$, the measured angular rate vector $\boldsymbol{\omega}$ and where $[\boldsymbol{\omega}\times]$ is a skew-symmetric matrix.
The dynamic state vector becomes
\begin{align}
\boldsymbol{x} &= \begin{bmatrix}
\boldsymbol{v}^\top & \boldsymbol{q}^\top&\boldsymbol{p}^\top\end{bmatrix}^\top
\end{align}
The wheel encoders are used to obtain the vehicle's longitudinal velocity $v_x$, assuming there is no wheel slip and the wheel radii are calibrated. For the vertical velocity $v_z$ we make a zero-mean assumption, thus neglecting the relative movement of the vehicle body against the road. To obtain a measurement of the lateral velocity $v_y$ we use our new $\omega_z$-$v_y$ model:
\begin{align}
v_y = (l_x-x_{\rho})\,\omega_z,
\end{align}
where $l_{x}$ describes the lever arm from the rear axis to the IMU. The zero-slip model uses the same measurement equation with $x_\rho=0$.
We apply an extended Kalman filter to this dynamic system. It is well known that the both the position and heading are unobservable, when only providing velocity measurements. Thus, these states remain unstabilized.

\subsection{Lateral Velocity Estimation}
\begin{figure}[ht]
\centering
\includegraphics[width=0.7\linewidth]{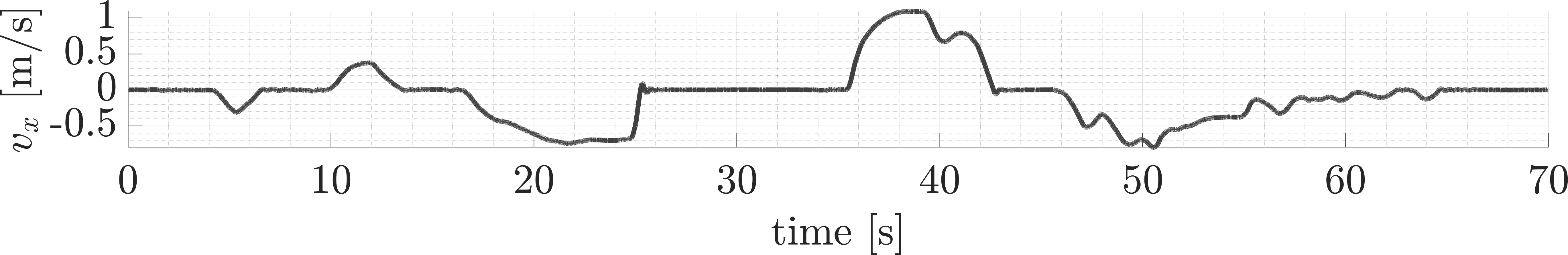}
\includegraphics[width=0.7\linewidth]{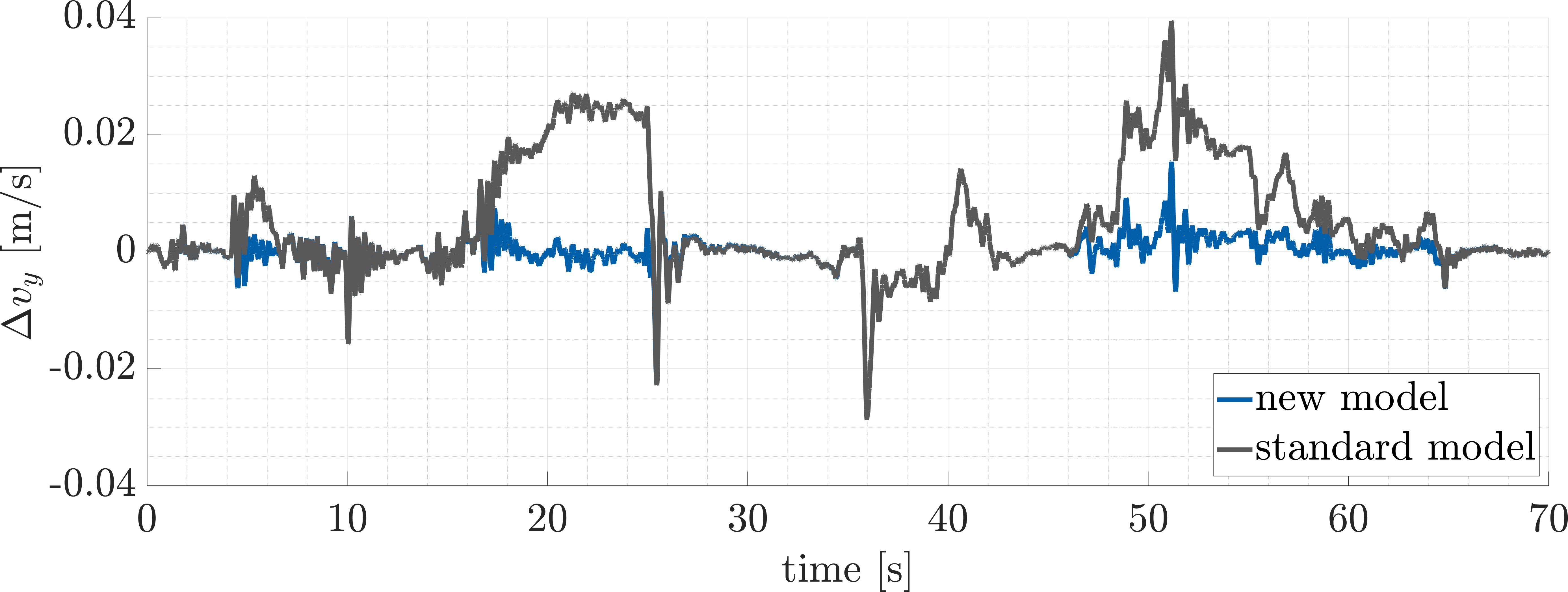}
\caption{Lateral velocity error $\Delta v_y$ for parking maneuvers, with new $\omega_z$-$v_y$ model (blue) compared against the zero-slip model (gray). Upper plot depicts the longitudinal velocity during the maneuver.}
\label{fig:vy_plot}
\end{figure}

We first evaluate the results on the estimation of the lateral velocity $v_y$. We performed several parking maneuvers at low speeds. Fig. \ref{fig:vy_plot} depicts one of those parking maneuvers with several corrections both in forward and reverse direction. We can see the zero-slip model estimating the lateral velocity with errors up to 2 cm/s. Whereas our new model captures the increase in lateral velocity more accurately. Especially for reverse driving the improvements are directly visible.

\begin{figure}[ht]
\centering
\includegraphics[width=0.4\linewidth]{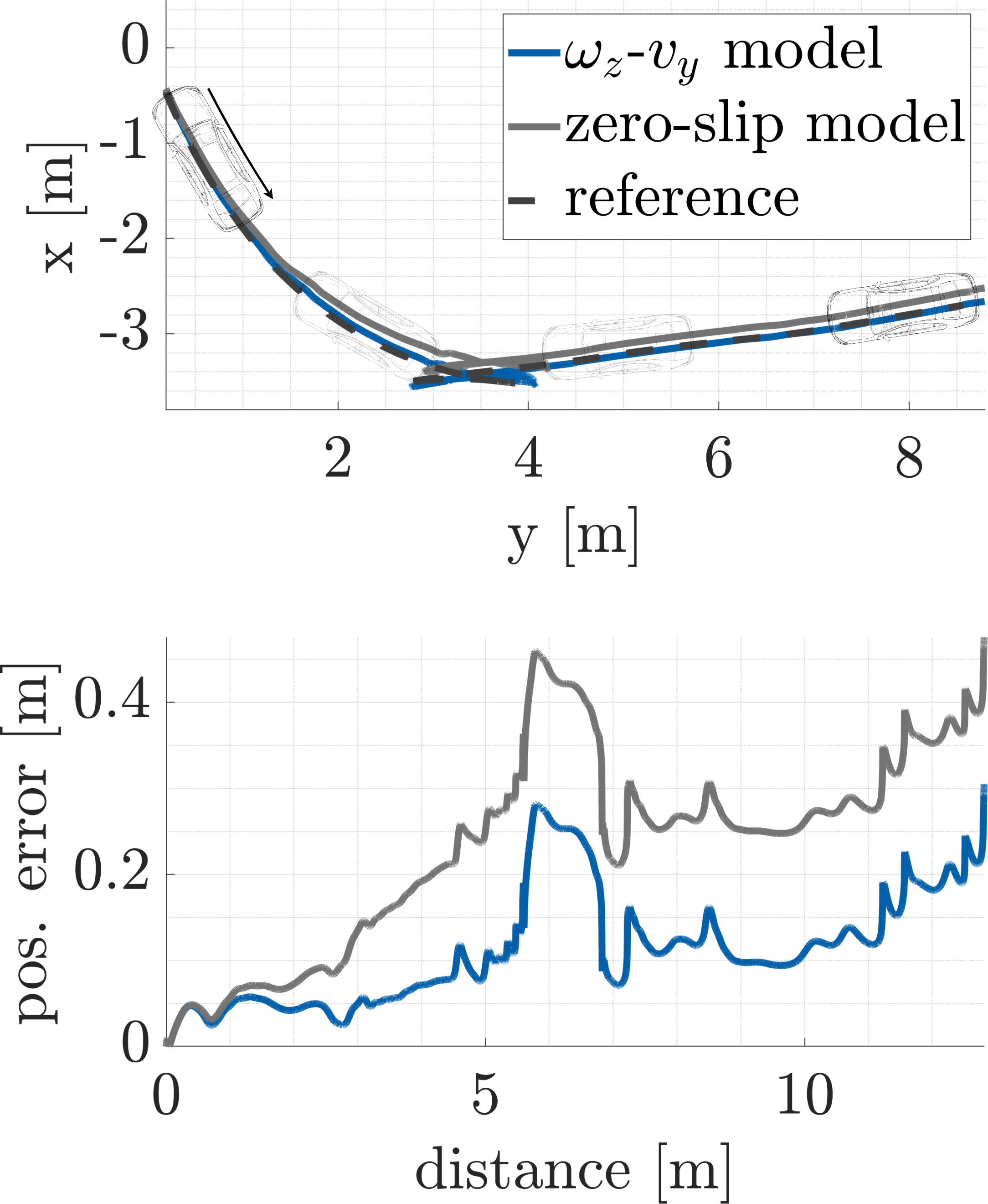}
\includegraphics[width=0.4\linewidth]{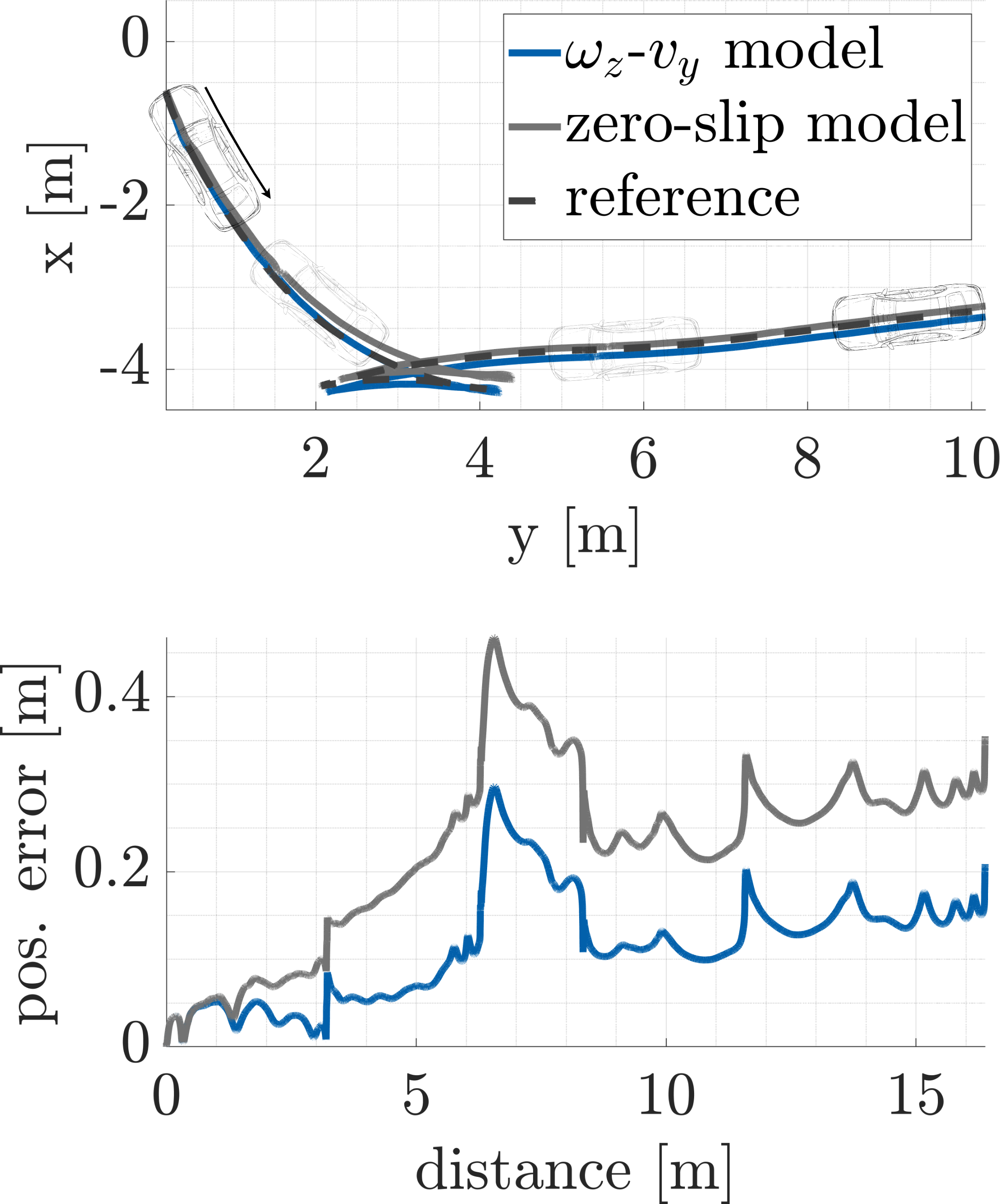}
\caption{Localization for two perpendicular parking maneuvers. The upper plots show the trajectories, while the plots below depict the error between the reference trajectory and the estimated trajectories. The remaining localization error is contributed to wheel-odometry errors.}
\label{fig:plot_03}
\end{figure}

\subsection{Parking Localization Accuracy}
We evaluate the impact of the proposed model on localization performance during parking maneuvers by analyzing the trajectory error, defined as the Euclidean distance between the estimated and true vehicle positions over the entire maneuver.

The initial data set consists of nine parking maneuvers, including both perpendicular and parallel maneuvers in forward and reverse directions. Fig.~\ref{fig:plot_03} illustrates two reverse perpendicular parking maneuvers, where the improved localization performance using the new model is clearly visible. As expected, the model has the greatest impact on reverse perpendicular maneuvers. We further evaluated the remaining localization error and found that it is caused by errors in the wheel-odometry measurement. If ideal longitudinal velocity measurements are used, the localization error is reduced to a few centimeters. The new lateral velocity model thus captures the lateral effects with sufficient accuracy.

\begin{figure}[ht]
\centering
\resizebox{0.8\linewidth}{!}{%
\begin{tikzpicture}
    \begin{axis}[
    	ybar,
    	stack negative=separate,
        width  = \linewidth,
        height = 5cm,
        major x tick style = transparent,
        ybar=5*\pgflinewidth,
        bar width=5pt,
        ymajorgrids = true,
        ylabel = {trajectory error [m]},
        xlabel = {Parking Maneuvers},
        symbolic x coords={1, 2, 3, 4, 5, 6, 7, 8, 9},
		xlabel near ticks,
		ylabel near ticks,
        scaled y ticks = false,
        xmajorticks=false,
        enlarge x limits=0.05,
        font=\footnotesize,
        axis lines*=left,
        point meta=explicit symbolic,
        legend pos=north west,
        %restrict y to domain*=0:0, % negative values
        %y=1pt,  % important workaround!
    ]
        \addplot[style={bblue,fill=bblue,mark=none}]
            coordinates {(1,0.105) (2,0.090) (3,0.159) (4,0.094) (5,0.092) (6,0.165) (7,0.130) (8,0.140) (9,0.134)};             
        \addlegendentry{$\omega_z$-$v_y$ model}
        
        \addplot[style={ggray,fill=ggray,mark=none}]
             coordinates {(1,0.130) (2,0.088) (3,0.187) (4,0.091) (5,0.117) (6,0.307) (7,0.249) (8,0.210) (9,0.144)};
        \addlegendentry{zero-slip model}
        \pgfplotsextra{
    \draw[dashed, thick, bblue]
        ([shift={(-0.3cm,0cm)}]axis cs:6,0)
        rectangle
        ([shift={(0.3cm,0.1cm)}]axis cs:6,0.307);
    \draw[dashed, thick, bblue]
        ([shift={(-0.3cm,0cm)}]axis cs:7,0)
        rectangle
        ([shift={(0.3cm,0.1cm)}]axis cs:7,0.249);
    \draw[dashed, thick, bblue]
        ([shift={(-0.3cm,0cm)}]axis cs:8,0)
        rectangle
        ([shift={(0.3cm,0.1cm)}]axis cs:8,0.210);
}
    \end{axis}
\end{tikzpicture}
}

\caption{Localization performance for individual parking maneuvers. The end-position error is the final localization error for each individual maneuver. The highlighted maneuvers are perpendicular-reverse-parking maneuvers, where the largest performance improvements are achieved.}
\label{fig:bar}
\end{figure}
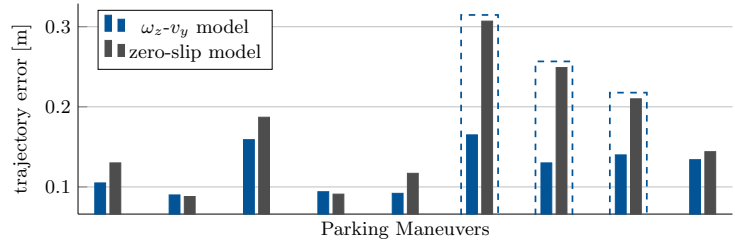

A summary of all nine maneuvers is presented in Fig.~\ref{fig:bar}, which shows the trajectory errors for each case. The most significant improvements are again observed for reverse perpendicular parking, highlighted with dotted boxes. In contrast, improvements are smaller for other scenarios, particularly for parallel parking maneuvers, where the influence of the model parameter is smaller, as previously demonstrated in the parameter disturbance analysis.

\renewcommand{\arraystretch}{1.3}
\begin{table}[ht]
\centering
\begin{threeparttable}
\caption{Evaluation of localization accuracy for perpendicular parking maneuvers.}
\label{tab:eval}
\begin{tabular}{wl{2.5cm}|wc{3cm}wc{3cm}wc{2.5cm}}
\hline
{\textbf{method}} & {\textbf{63rd percentile}} & \textbf{95th percentile} & {\textbf{maximum}}\\
\hline
{$\omega_z$-$v_y$ model} & \textbf{0.12 m} & \textbf{0.18 m} & \textbf{0.25 m}\\
\hline
{zero-slip model} & 0.15 m & 0.30 m & 0.38 m\\
\end{tabular}
\begin{tablenotes}
\item Note: The values for the 63rd and the 95th percentile indicate that 63\% or 95\% of all samples have an error of less than or equal to the value provided in the table. Depicted is the trajectory error over the whole maneuver.
\end{tablenotes}
\end{threeparttable}
\end{table}

We evaluate a second dataset consisting exclusively of perpendicular parking maneuvers. While these maneuvers are simpler in terms of duration and execution, the increased sample size (45 maneuvers in total) enables a more robust statistical analysis. Table~\ref{tab:eval} summarizes the trajectory errors across all 45 maneuvers. In 63\% of the test cases (performance threshold), the proposed model and the zero-slip model achieve errors at or below 15 cm, with the proposed model showing a slight performance advantage.

In terms of consistency, reflected by the 95th percentile error, our model maintains a position error within 18 cm, compared to 30 cm for the zero-slip model. The maximum observed error is 25 cm for the proposed model and 38 cm for the zero-slip approach. These results demonstrate that our simple model improves localization accuracy for parking odometry, without relying on complex model assumptions or heuristic adjustments.

\section{Conclusion}
We introduced a lightweight lateral velocity model for parking localization algorithms. The model distinguishes between forward and reverse driving by incorporating two separate parameters, which can be identified using high-precision ground-truth measurements. In support of the model, we provided physical insights into the underlying vehicle dynamics. Additionally, we conducted a parameter disturbance analysis, quantifying how parameter deviations translate into localization errors. This analysis offers a practical foundation for deriving parameter requirements. Experiments showed increased localization accuracy when using the model for parking localization. 

Future work will explore online estimation of the model parameters using automotive-grade sensors. This way, no additional measurement equipment is necessary to fit the model. Furthermore, the transition from low-speed parking to normal driving could be investigated, as the discussed low-speed effects lose influence with increasing speed and, thus, regular tire-force models become more applicable.

\section*{Acknowledgment}
This work is a result of the joint research project STADT:up (19A22006x).
The project is supported by the German Federal Ministry for
Economic Affairs and Climate Action, based on a decision of
the German Parliament.

\section*{Conflict of Interest}
The authors report no conflict of interest.

\bibliographystyle{abbrv}
\bibliography{bib/literatur}

\begin{thebibliography}{10}

\bibitem{Bai.2012}
F.~Bai, K.~H. Guo, B.~J. Zhang, and D.~Lu.
\newblock Analysis of {{Tire Cornering Stiffness Property}} under {{Slight
  Driving}} or {{Braking Condition}}.
\newblock {\em Applied Mechanics and Materials}, 271--272:767--772, Dec. 2012.

\bibitem{Besselink.2022}
I.~J.~M. Besselink, M.~H.~M. Baart, and H.~Nijmeijer.
\newblock Simplified turn slip modeling by a parallel magic formula model.
\newblock In A.~Orlova and D.~Cole, editors, {\em Advances in Dynamics of
  Vehicles on Roads and Tracks {{II}}}, pages 966--973, Cham, 2022. Springer
  International Publishing.

\bibitem{Bev.2016}
M.~Bevilacqua, A.~Tsourdos, and A.~Starr.
\newblock Egomotion estimation for monocular camera visual odometer.
\newblock In {\em 2016 {{IEEE}} International Instrumentation and Measurement
  Technology Conference Proceedings}, pages 1--6, 2016.

\bibitem{Brossard.2019}
M.~Brossard, A.~Barrau, and S.~Bonnabel.
\newblock {{AI-IMU Dead-Reckoning}}, Apr. 2019.

\bibitem{Brunker.2019}
A.~Brunker, T.~Wohlgemuth, M.~Frey, and F.~Gauterin.
\newblock Odometry 2.0: {{A Slip-Adaptive EIF-Based Four-Wheel-Odometry Model}}
  for {{Parking}}.
\newblock {\em T-IV}, 4(1):114--126, Mar. 2019.

\bibitem{Caltabiano.2004}
D.~Caltabiano, G.~Muscato, and F.~Russo.
\newblock Localization and self-calibration of a robot for volcano exploration.
\newblock In {\em {{IEEE}} International Conference on Robotics and Automation,
  2004. {{Proceedings}}. {{ICRA}} '04. 2004}, volume~1, pages 586--591 Vol.1,
  2004.

\bibitem{Cao.2019}
D.~Cao, B.~Tang, H.~Jiang, C.~Yin, D.~Zhang, and Y.~Huang.
\newblock Study on {{Low-Speed Steering Resistance Torque}} of {{Vehicles
  Considering Friction}} between {{Tire}} and {{Pavement}}.
\newblock {\em Applied Sciences}, 9(5):1015, Mar. 2019.

\bibitem{Diener.2025}
L.~Diener, J.~Kalkkuhl, and M.~Enzweiler.
\newblock Radar {{Misalignment Calibration}} using {{Vehicle Dynamics Model}}.
\newblock {\em TechRxiv preprint}, Feb. 2025.

\bibitem{Diener.2024}
L.~Diener, J.~Kalkkuhl, and T.~Schirle.
\newblock Radar-based approach for side-slip gradient estimation.
\newblock {\em SAE Technical Paper 2024-01-2976}, July 2024.

\bibitem{Du.2023}
B.~Du, H.~Wang, S.~Pan, D.~Liu, Y.~Zhu, and Z.~Shi.
\newblock Robust {{Multilayer Vehicle Model-Aided INS Based}} on {{Soft}} and
  {{Hard Constraintsz}}.
\newblock {\em IEEE Sensors Journal}, 23(1):812--827, Jan. 2023.

\bibitem{Fazekas.2020}
M.~Fazekas, B.~N{\'e}meth, and P.~G{\'a}sp{\'a}r.
\newblock Model {{Based Vehicle Localization}} via an {{Iterative Parameter
  Estimation}}.
\newblock In M.~Klomp, F.~Bruzelius, J.~Nielsen, and A.~Hillemyr, editors, {\em
  Advances in {{Dynamics}} of {{Vehicles}} on {{Roads}} and {{Tracks}}}, pages
  1660--1669. Springer International Publishing, Cham, 2020.

\bibitem{Gallrein.2014}
A.~Gallrein, M.~Baecker, M.~Burger, and A.~Gizatullin.
\newblock An advanced flexible realtime tire model and its integration into
  fraunhofer's driving simulator.
\newblock In {\em {{SAE}} 2014 World Congress \& Exhibition}. SAE
  International, Apr. 2014.

\bibitem{Garcia.2014}
D.~{Garcia-Pozuelo}, V.~Diaz, and M.~J.~L. Boada.
\newblock New tyre-road contact model for applications at low speed.
\newblock {\em International Journal of Automotive Technology}, 15(4):553--564,
  June 2014.

\bibitem{Guo.2005}
K.~Guo, Y.~Zhuang, D.~Lu, S.-k. Chen, and W.~Lin.
\newblock A study on speed-dependent tyre--road friction and its effect on the
  force and the moment.
\newblock {\em Vehicle System Dynamics}, 43(sup1):329--340, Jan. 2005.

\bibitem{Han.2020}
C.~Han, M.~Frey, and F.~Gauterin.
\newblock Modular {{Approach}} for {{Odometry Localization Method}} for
  {{Vehicles}} with {{Increased Maneuverability}}.
\newblock {\em Sensors}, 21(1):79, Dec. 2020.

\bibitem{Han.2025}
C.~Han, M.~Frey, and F.~Gauterin.
\newblock A {{Simulation-Based Efficient Optimization Method}} of an {{Odometry
  Localization Filter}} for {{Vehicles}} with {{Increased Maneuverability}}.
\newblock {\em OJ-ITS}, 2025.

\bibitem{Klier.2008}
W.~Klier, A.~Reim, and D.~Stapel.
\newblock Robust estimation of vehicle sideslip angle - an approach w/o vehicle
  and tire models.
\newblock In {\em {{SAE}} World Congress \& Exhibition}. SAE International,
  Apr. 2008.

\bibitem{Kochem.2002}
M.~Kochem, R.~Neddenriep, R.~Isermann, N.~Wagner, and C.-D. Hamann.
\newblock Accurate local vehicle dead-reckoning for a parking assistance
  system.
\newblock In {\em Proceedings of the 2002 {{American Control Conference}}
  ({{IEEE Cat}}. {{No}}.{{CH37301}})}, pages 4297--4302 vol.5, Anchorage, AK,
  USA, 2002. IEEE.

\bibitem{Liang.2022}
Y.~Liang, S.~M{\"u}ller, and D.~Rolle.
\newblock Tightly coupled multimodal sensor data fusion for robust state
  observation with online delay estimation and compensation.
\newblock {\em IEEE Sensors Journal}, 22(13):13480--13496, 2022.

\bibitem{Lugaro.2016}
C.~Lugaro, A.~Schmeitz, T.~Ogawa, T.~Murakami, and S.~Huisman.
\newblock Development of a {{Parameter Identification Method}} for
  {{MF-Tyre}}/{{MF-Swift Applied}} to {{Parking}} and {{Low Speed Manoeuvres}}.
\newblock {\em SAE International Journal of Passenger Cars - Mechanical
  Systems}, 9(2):892--902, Apr. 2016.

\bibitem{Mori.2025}
D.~Mori, M.~Kamekawa, N.~Fujieda, and Y.~Mizuno.
\newblock Robust {{Vehicle State Estimation From Urban Driving}} to the
  {{Limits}} of {{Handling}}.
\newblock {\em IEEE Transactions on Intelligent Transportation Systems},
  26(1):892--905, Jan. 2025.

\bibitem{Oertel.2012}
C.~Oertel and Y.~Wei.
\newblock Tyre rolling kinematics and prediction of tyre forces and moments:
  Part {{I}} -- theory and method.
\newblock {\em Vehicle System Dynamics}, 50(11):1673--1687, Nov. 2012.

\bibitem{Pacejka.2012}
H.~B. Pacejka and I.~Besselink.
\newblock {\em Tire and Vehicle Dynamics}.
\newblock Butterworth-Heinemann Elsevier, Oxford Waltham, 3d edition edition,
  2012.

\bibitem{Marco.2020}
V.~Rodrigo~Marco, J.~Kalkkuhl, J.~Raisch, W.~J. Scholte, H.~Nijmeijer, and
  T.~Seel.
\newblock Multi-modal sensor fusion for highly accurate vehicle motion state
  estimation.
\newblock {\em CEP}, 100:104409, 2020.

\bibitem{Simione.2002}
P.~Simionescu and D.~Beale.
\newblock Optimum synthesis of the four-bar function generator in its symmetric
  embodiment: The {{Ackermann}} steering linkage.
\newblock {\em Mechanism and Machine Theory}, 37(12):1487--1504, Dec. 2002.

\bibitem{Sola.2017}
J.~Sol{\`a}.
\newblock Quaternion kinematics for the error-state {{Kalman}} filter.
\newblock {\em preprint}, Nov. 2017.

\bibitem{Vietinghoff.2007}
A.~{von Vietinghoff}, S.~Olbrich, and U.~Kiencke.
\newblock Extended {{Kalman}} filter for vehicle dynamics determination based
  on a nonlinear model combing longitudinal and lateral dynamics.
\newblock In {\em {{SAE}} World Congress 2007, Detroit, April 16-19, 2007},
  volume 2007-01-0834 of {\em {{SAE}} Paper}, pages 1--8, 2007.

\bibitem{Wei.2016}
Y.~Wei, C.~Oertel, Y.~Liu, and X.~Li.
\newblock A theoretical model of speed-dependent steering torque for rolling
  tyres.
\newblock {\em Vehicle System Dynamics}, 54(4):463--473, Apr. 2016.

\end{thebibliography}

\end{document}